\documentclass[10pt,twocolumn,letterpaper]{article}
\pdfoutput=1
\usepackage{cvpr}              % To produce the CAMERA-READY version

\makeatletter
\@namedef{ver@everyshi.sty}{}
\makeatother

\usepackage{graphicx}
\usepackage{amsmath}
\usepackage{amssymb}
\usepackage{booktabs}

\usepackage[pagebackref,breaklinks,colorlinks]{hyperref}

\usepackage[capitalize]{cleveref}
\crefname{section}{Sec.}{Secs.}
\Crefname{section}{Section}{Sections}
\Crefname{table}{Table}{Tables}
\crefname{table}{Tab.}{Tabs.}

\usepackage[table,xcdraw]{xcolor}
\usepackage{amsmath}
\usepackage{epsfig}
\usepackage{graphics}
\usepackage{multirow}
\usepackage{subcaption}
\usepackage{mwe}
\usepackage{makecell}
\usepackage{url}
\usepackage{soul}

\newcommand{\comment}[1]{}

\usepackage[accsupp]{axessibility} % Improves PDF readability for those with disabilities.

\usepackage[normalem]{ulem} %strikethrough

\usepackage{xspace}

\usepackage{hyperref} %Used for \autoref
\usepackage{bm} %boldmath
\usepackage{amsmath} %text in equation
\usepackage{amssymb}
\usepackage{cleveref}

\usepackage{adjustbox} %table stuff
\usepackage{multirow}
\usepackage[para,online,flushleft]{threeparttable}
\usepackage{booktabs}
\usepackage[]{tikz}
\usepackage{algorithmicx}
\usepackage{algorithm}
\usepackage[noend]{algpseudocode}

\usepackage{color}

\newcommand{\equaref}[1]{ Eq. \ref{#1}}
\newcommand{\figref}[1]{ Fig. \ref{#1}}
\newcommand{\tabref}[1]{ Table \ref{#1}}

\let\oldtextbf\textbf
\renewcommand{\textbf}[1]{\oldtextbf{\boldmath #1}}

\usepackage{graphicx}
\usepackage{tikz}

\tikzset{
    path image/.style 2 args={
        path picture={
            \node at (path picture bounding box.center) {
                \includegraphics[height=#1]{#2}
            };
        }
    },
    draw/.style = {white}
}

\usepackage{float}

\def\cvprPaperID{30} % *** Enter the EarthVision Paper ID here (taken from CMT)
\def\confName{EarthVision}
\def\confYear{2023}

\begin{document}

%%%%%%%%% TITLE - PLEASE UPDATE
\title{UnCRtainTS: \underline{Uncertainty} Quantification for \underline{C}loud \underline{R}emoval \\ in Optical Satellite \underline{T}ime \underline{S}eries}

\author{Patrick Ebel$^{*}$\\
{\tt\small patrick.ebel@tum.de}
% For a paper whose authors are all at the same institution,
% omit the following lines up until the closing ``}''.
% Additional authors and addresses can be added with ``\and'',
% just like the second author.
% To save space, use either the email address or home page, not both
\and
Vivien Sainte Fare Garnot$^{\dag}$\\
{\tt\small vsaint@ics.uzh.ch}
\and
Michael Schmitt$^{\ddag}$\\
{\tt\small michael.schmitt@unibw.de}
\and
Jan Dirk Wegner$^{\dag}$\\
{\tt\small jandirk.wegner@uzh.ch}
\and
Xiao Xiang Zhu$^{*}$\\
{\tt\small xiaoxiang.zhu@tum.de}
\and
{\small $^{*}$ Technical University of Munich\quad $^{\dag}$ University of Zurich \quad  $^{\ddag}$ University~of~the~Bundeswehr~Munich
}
}

\maketitle

% As a general rule, do not put math, special symbols or citations
% in the abstract or keywords.
%\begin{abstract}
%The presence of clouds and haze prevents an ongoing monitoring of Earth via optical satellites. Preceding work on cloud removal restores obscured pixels with their most likely values, as based on spatio-spectral context, multi-sensory or multi-temporal data. So far, it is an open question how reliable reconstructed optical satellite images are. Our work introduces a novel attention-based model for cloud removal, termed UnCRtainTS, integrating all of the aforementioned sources of information and providing calibrated heteroscedastic aleatoric uncertainty estimates for its reconstructions. The proposed model performs best on two established benchmarks and is analysed in terms of its goodness of restorations as well as calibration.
%\end{abstract}

\begin{abstract}
Clouds and haze often occlude optical satellite images, hindering continuous, dense monitoring of the Earth's surface. Although modern deep learning methods can implicitly learn to ignore such occlusions, explicit cloud removal as pre-processing enables manual interpretation and allows training models when only few annotations are available.
Cloud removal is challenging due to the wide range of occlusion scenarios---from scenes partially visible through haze, to completely opaque cloud coverage. Furthermore, integrating reconstructed images in downstream applications would greatly benefit from trustworthy quality assessment. 
In this paper, we introduce UnCRtainTS, a method for multi-temporal cloud removal combining a novel attention-based architecture, and a formulation for multivariate uncertainty prediction. These two components combined set a new state-of-the-art performance in terms of image reconstruction on two public cloud removal datasets. Additionally, we show how the well-calibrated predicted uncertainties enable a precise control of the reconstruction quality.
\end{abstract}

% Note that keywords are not normally used for peerreview papers.
%\begin{IEEEkeywords}
%image reconstruction, multi-modal, multi-temporal, uncertainty.
%\end{IEEEkeywords}

% For peer review papers, you can put extra information on the cover
% page as needed:
% \ifCLASSOPTIONpeerreview
% \begin{center} \bfseries EDICS Category: 3-BBND \end{center}
% \fi
%
% For peerreview papers, this IEEEtran command inserts a page break and
% creates the second title. It will be ignored for other modes.
%\IEEEpeerreviewmaketitle

\section{Introduction}

\begin{figure}[ht!]
    \centering
    \includegraphics[width=.8\linewidth, trim=0cm 0cm 8cm 1cm, clip]{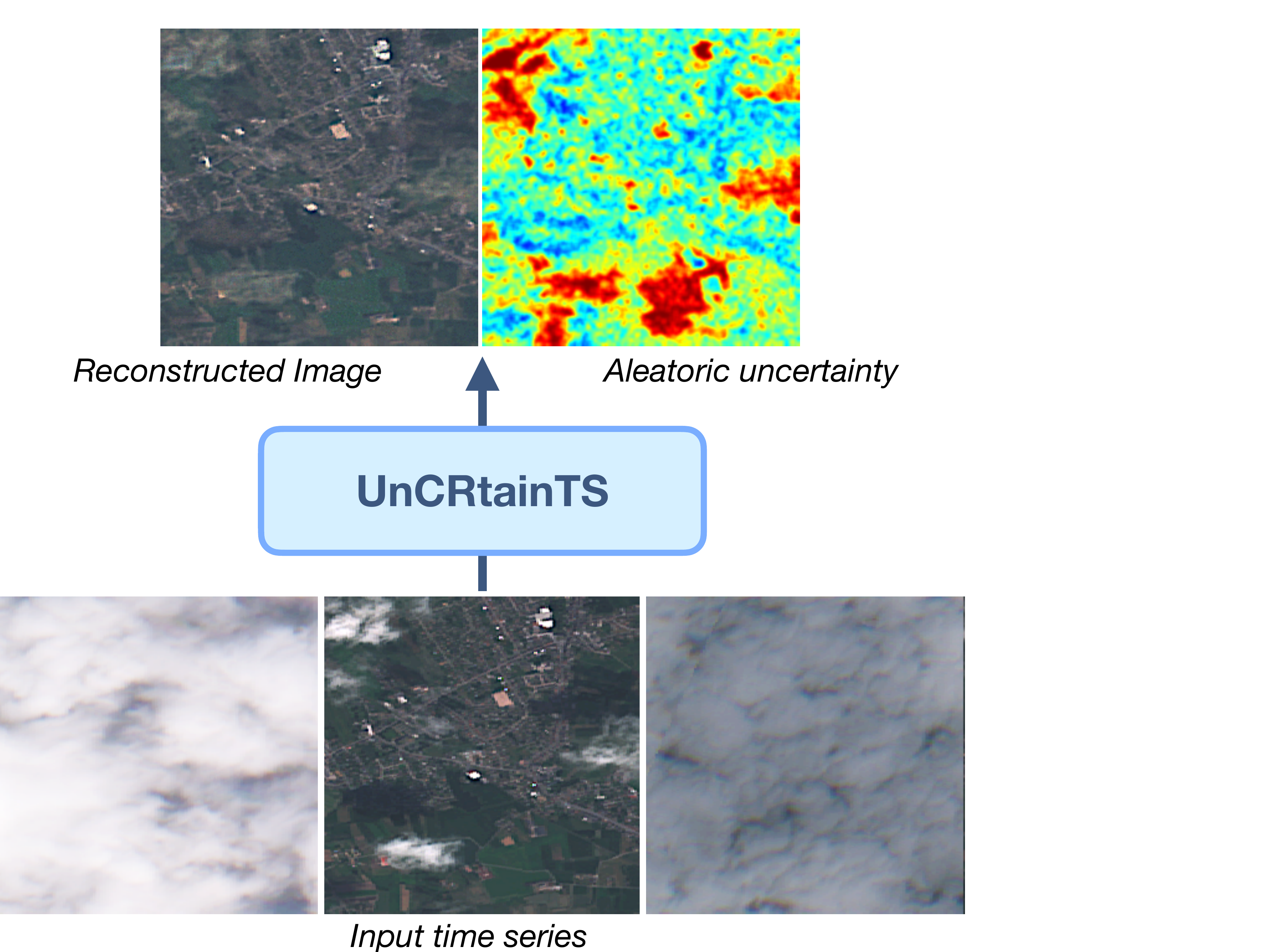}
    \caption{\textbf{Overview:} Our attention-based UnCRtainTS architecture predicts a single cloud-free image from a sequence of cloudy observations. For each reconstructed pixel, our method also estimates the aleatoric uncertainty of the prediction. Note how higher uncertainties (in red) are associated with persistent occlusion, cloud shadow, or with specific land cover types. 
    }
    \label{fig:teaser}
\end{figure}

Multispectral, optical satellite imagery allows for large-scale assessments of the environment like crop monitoring \cite{russwurm2017temporal,TURKOGLU2021112603} and global vegetation height estimation \cite{lang2022high,lang2022global}. Clouds, haze and other atmospheric disturbances, however, often occlude large parts of optical satellite images, 
particularly during meteorological winter season \cite{King_Platnick_Menzel_Ackerman_Hubanks_2013} and over landcover such as rainforests \cite{Bermudez_Happ_Oliveira_Feitosa_2018}. Neural networks trained on extensive amounts of annotated data may implicitly learn to ignore task-irrelevant cloudy observations \cite{russwurm2020self, russwurm2017temporal,metzger2021}. Yet, explicit cloud removal as a pre-processing step can further improve model performance and is valuable if ground truth annotations for supervised training are scarce \cite{explicitCR}. Cloud removal prior to training or applying a pre-trained task-specific model also permits a seamless analysis using traditional non-learning methods or visualisation \cite{LIU2021112364}.

\noindent

Hence,  cloud removal is an active field of research boasting a large body of literature on image reconstruction methods to recover cloud-free observations \cite{Enomoto_Sakurada_Wang_Fukui_Matsuoka_Nakamura_Kawaguchi_2017, Bermudez_Happ_Oliveira_Feitosa_2018, grohnfeldt2018conditional, meraner2020cloud, ebel2020multisensor, gao2020cloud, Sarukkai_Jain_Uzkent_Ermon_2019,  sebastianelli2022plfm}. Such methods are typically evaluated in terms of image restoration metrics, e.g. mean squared error or structural similarity (SSIM), providing an aggregated measure of reconstruction quality. These metrics, however, provide little insight into how reliable a given reconstruction is on a pixel-wise or image-by-image basis. To address this shortcoming, we introduce uncertainty estimation to satellite image reconstruction, specifically to the task of multi-temporal cloud-removal in optical satellite images.
 Predicting uncertainties that correlate with the empirical errors of a neural net is at the core of the growing field of probabilistic deep learning \cite{kendall2017uncertainties, takahashi2018student, skafte2019reliable}. 
 By modelling the uncertainty and training for a negative log likelihood (NLL) objective, such approaches allow to jointly learn a model for making a prediction and estimate the prediction's variances. 
 If well-calibrated, the predicted uncertainties can be very valuable for downstream usage by providing a measure of a reconstruction's confidence. 
Uncertainty quantification has been successfully applied in univariate remote sensing regression problems such as canopy height regression \cite{lang2022global} or flood risk estimation \cite{chaudhary2022flood}. 
Here, we extend uncertainty quantification to multivariate regression for satellite image reconstruction. We obtain experimentally well-calibrated uncertainties that enable flagging poorly reconstructed images. We also show that multivariate uncertainty prediction requires a multivariate uncertainty model for better calibration.

Aleatoric uncertainty prediction implies training with a pixel-based Negative Log Likelihod (NLL) loss.
On the other hand, image reconstruction losses like SSIM or perceptual loss are typically used in existing cloud removal methods to better retrieve high-frequency details \cite{ebel2020multisensor, wang2022hybrid, darbaghshahi22}. Here, we introduce a novel neural architecture that operates on feature maps at full resolution. It leverages attention-based temporal encoding, allowing it to outperform previous state-of-the-art approaches even when trained via a pixel-based loss.
In sum, our contributions are: 
\begin{itemize}
    \item We introduce multivariate uncertainty quantification to the task of multispectral satellite image reconstruction, to obtain both reconstructions and variance estimates. 
    \item We propose a novel neural network architecture achieving state-of-the-art results on two challenging benchmark datasets for optical satellite cloud removal.
    \item We obtain well-calibrated uncertainties that allow to measure and control the quality of reconstructed images for risk-mitigation in downstream applications.
\end{itemize}

\section{Related Work}
\subsection{Cloud Removal in Satellite Image Time Series}
\label{sec:related:CR}
Optical satellite image reconstruction \cite{shen2015missing}, and specifically cloud removal, pose a long-standing challenge in remote sensing \cite{lin2012cloud, eckardt2013removal, li2014recovering, hu2015thin, huang2015sar}. Contemporary deep learning approaches can be categorised into mono-temporal \cite{Enomoto_Sakurada_Wang_Fukui_Matsuoka_Nakamura_Kawaguchi_2017, Bermudez_Happ_Oliveira_Feitosa_2018, gao2020cloud, pan2020cloud, xu2022glf}, mono-temporal \& multi-modal \cite{grohnfeldt2018conditional, meraner2020cloud, ebel2020multisensor}, multi-temporal \cite{Sarukkai_Jain_Uzkent_Ermon_2019} and multi-temporal \& multi-modal methods \cite{ebel2022sen12ms, sebastianelli2022plfm}. Here, we consider the reconstruction task in a multi-temporal \& multi-modal setting.

Spatial encoding of image reconstruction is either done with UNet-like encoder-decoder backbones \cite{ronneberger2015u, isola2017image, zhu2017unpaired} that spatially down-sample the intermediate representations\cite{Enomoto_Sakurada_Wang_Fukui_Matsuoka_Nakamura_Kawaguchi_2017,grohnfeldt2018conditional, ebel2020multisensor}, or with architectures preserving the full resolution of the images \cite{lanaras2018super, meraner2020cloud}. While the first are computationally more efficient especially in the multi-temporal setting, the latter tend to better preserve the spatial structure in the reconstructed images. In fact, downsampling architectures often necessitate auxiliary perceptual \cite{johnson2016perceptual, ebel2020multisensor, hwang2020sar, ebel2021internal} or structural similarity losses \cite{Wang_Bovik_Sheikh_Simoncelli_2004, thin_thick_cr} to recover high-frequency information. The combination of such cost functions with a probabilistic training objective for uncertainty prediction is not straightforward. Therefore, we design an architecture that operates on full resolution feature maps and make design choices to reduces its computational complexity.
For temporal encoding, we draw inspiration from recent work in satellite time series encoding \cite{garnot2020lightweight, garnot2021panoptic, russwurm2020self} and rely on self-attention to integrate the temporal information.

\subsection{Uncertainty Quantification}
\label{sec:related:uncertainty}

Uncertainty can be partitioned into \emph{epistemic} or model uncertainty, and \emph{aleatoric} or data uncertainty. Epistemic uncertainty accounts for the uncertainty on the model's weights, and can be estimated for instance with ensemble methods \cite{lakshminarayanan2017simple, turkoglu2022filmensemble}, or monte-carlo dropout \cite{gal2016dropout} in deep nets. 
Aleatoric uncertainty captures the randomness inherent to the data. In the case of optical satellite image reconstruction, aleatoric uncertainty may thus help flagging restorations based on too little evidence. 
In the recent deep learning literature, aleatoric uncertainty estimation is achieved via likelihood maximization with a parametric model of the noise distribution \cite{takahashi2018student, skafte2019reliable, seitzer2021pitfalls, ansariautoinverse, stirn2022faithful}. This is a common technique in safety-critical applications, such as solving inverse problems in biomedical imaging \cite{bhadra2021hallucinations, edupuganti2020uncertainty, laves2020uncertainty, laves2020well, tolle2021mean, antoran2022bayesdip, chung2022mr, glaubitz2023generalized}.
Uncertainty quantification is of growing interest in remote sensing \cite{gawlikowski2021survey}, with applications to forest assessments, flood hazard monitoring, geophysical modeling, landcover classification and out-of-distribution detection \cite{lang2022global, lang2022high, chaudhary2022flood, liu2022uncertainty, gawlikowski2022robust, gawlikowski2022advanced}. 
As prior remote sensing work covers uncertainty quantification for univariate regression problems, the multivariate extension has yet to be explored. 
To our knowledge, the aforementioned contributions are either on image reconstruction in the biomedical domain or target specific remote sensing downstream tasks, such that ours is the first work to investigate uncertainty quantification for multispectral satellite image reconstruction. 
The current lack of uncertainty quantification in the cloud removal literature is a significant research gap because reconstructed satellite images may guide safety-critical downstream applications or human judgement alike, such that pixel-wise measures of confidence would be beneficial.

\begin{figure*}[ht!]
    \includegraphics[width=\linewidth, trim=0cm 21cm 1cm 0cm, clip]{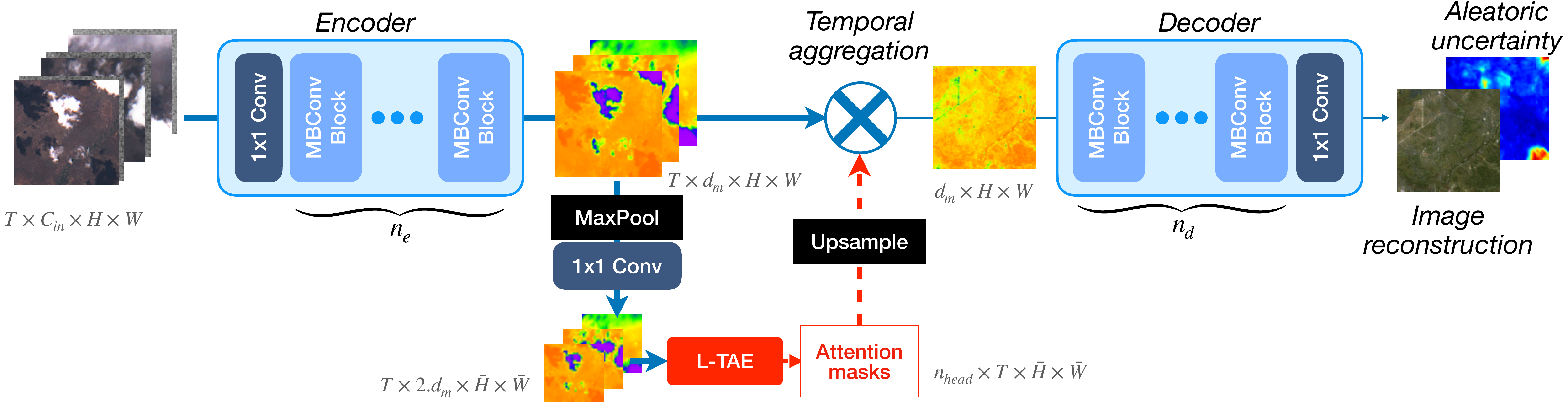}
    \caption{ \textbf{UnCRtainTS.}
    The network consists of three main parts, applied along a main branch of 
    MBConv blocks \cite{sandler2018mobilenetv2} that is processing feature maps at full input resolution: First, an \textit{encoder} is applied in parallel to the $T$ time points. Then, an \textit{attention-based temporal aggregator} computes attention mask by applying an L-TAE to downsampled feature maps, used to aggregate the sequence of observations. Finally, the temporally integrated feature map is processed by a \textit{decoding block}, yielding the image reconstruction and aleatoric uncertainty.
    }
    \label{fig:architecture_uncrtaints}
\end{figure*}

\section{Methods}

We follow the problem statement of the public cloud removal benchmark SEN12MS-CR-TS \cite{ebel2022sen12ms}. Each sample $i$ of the $N$-sized dataset consists of a pair $(\boldsymbol{X^i},Y^i)$, where $\boldsymbol{X}^i=[X^i_1, \cdots, X^i_T]$ is the input time series of size $\left[T\times C_{in} \times H \times W \right]$ containing cloudy pixels, and $Y^i$ is the target cloud-free image of shape $\left[ K \times H \times W \right]$. $T$ denotes the number of dates in the input sequence, $C_{in}$ and $K$ the number of input and output channels, and $H\times W$ the two spatial dimensions of the images.  As in \cite{ebel2022sen12ms}, we set $T=3$, $C_{in}=15$, $K=13$, $H=W=256$. Note that $C_{in} \neq K$ because Sentinel-1 radar observations are utilized as additional input. Furthermore, aleatoric uncertainty quantification introduces additional output channels to describe the modeled noise distribution. For convenience, we drop the $i$ superscript in the rest of this section.

\subsection{Network Architecture}
Our proposed UnCRtainTS network architecture maps a cloudy input time series to a single cloud-free optical image. As explained in Sec. \ref{sec:related:CR}, we make the explicit choice to perform spatial encoding only on full-resolution feature maps to allow for good performance when training with a pixel-based loss. To ease the impact of this choice on the computational load of the architecture, we rely on efficient MBConv blocks \cite{sandler2018mobilenetv2}. They combine depthwhise convolution and regular pointwise convolutions for computationally efficient spatial encoding. We perform temporal encoding on downsampled feature maps via the attention-based L-TAE \cite{garnot2020lightweight},
which is designed for satellite image time series and computationally more efficient than transformers.
The network architecture is illustrated in \figref{fig:architecture_uncrtaints} and further described in the following paragraphs.

\paragraph{\bf Pre-aggregation shared encoder}
The $T$ different input images are processed in parallel by a shared spatial encoding branch. This encoder is composed of a pointwise convolution $C_{in} \rightarrow d_m$, followed by a specifiable number $n_e$ of MBConv blocks. Following \cite{garnot2021panoptic} we use group normalisation in the encoding branch.
All MBConv blocks map to $d_m \rightarrow 2 \times d_m \rightarrow d_m$ channels 
and contain Squeeze-Excitation layers \cite{hu2018squeeze}. Ultimately, each input image $X_{t}$ is mapped to a feature map $f_{t}$ of the same resolution.

\paragraph{\bf Attention-based temporal aggregation } Following recent literature, we employ self-attention to aggregate a sequence of feature maps $[f_1, \cdots, f_T]$ into a single one. We first down-sample features $f_t$ with a single max-pooling operation to low resolution feature maps $\overline{f_t}$ of size $\left[d_m \times \overline{H} \times \overline{W} \right]$. We set $\overline{H} = \overline{W} = 32$, to limit computation while providing sufficient resolution to group cloudy pixels, which typically cluster in space.  We re-project the downsampled features via a linear layer $d_m \rightarrow 2 \times d_m $.
Next, as in \cite{garnot2021panoptic}, the low-resolution features $\overline{f_t}$ are processed pixel-wise with an L-TAE \cite{garnot2020lightweight,garnot2020satellite}: we obtain attention masks over the $T$ observations  for each pixel position of the low resolution feature maps. Contrary to previous work, we only use the L-TAE's  attention masks, and omit attention-weighting of the sequence of low resolution feature maps. We upsample the attention masks to the full resolution via bilinear interpolation, and apply them to the sequence of high resolution feature maps $[f_1, \cdots, f_T]$. This results in a single feature map $\hat{f}$ of shape $\left[d_m \times H \times W \right]$.
We use a dropout rate of $0.1$ on the attention masks after upsampling, and the temporal aggregation is done with L-TAE's channel grouping strategy \cite{garnot2020lightweight}. 

\paragraph{\bf Post-aggregation decoding} The temporally aggregated feature map $\hat{f}$ is processed by a decoding branch, which consists of a specifiable number $n_d$ of batch-normalized MBConv blocks and a final $d_m \rightarrow C_{out}$ pointwise convolution followed by a non-linearity. 
For every channel predicting image reconstruction, we use a sigmoidal function to  squash the outputs into the data's valid range.
For channels predicting aleatoric uncertainty (see next section), we use a softplus activation to ensure positivity, as in \cite{seitzer2021pitfalls, harakeh2022estimating, stirn2022faithful}.

\subsection{Aleatoric uncertainty prediction}
\label{sub:optimization}

Here, we explain how our UnCRtainTS method predicts an aleatoric uncertainty value for each reconstructed pixel. As UnCRtainTS is trained with pixel-wise losses, we henceforth adopt a pixel-based notation. We consider the set of pixels of cardinal $n$ contained in the dataset. We denote each pixel reconstruction by $\boldsymbol{\hat{y}}_j$ and the corresponding ground truth by $\boldsymbol{y}_j$, both vectors of dimension $K$. 

\paragraph{\bf Image reconstruction} In the default setting of satellite image reconstruction, the network only regresses the target pixel values. Hence, in this setting, $C_{out}=K$ and the predictions are typically  supervised with L2 loss\cite{dong2015image,anwar2020deep}:

\begin{equation}
\mathcal{L}_2(\boldsymbol{\hat{y}},\boldsymbol{y}) = \frac{1}{n} \sum_{j=1}^{n} \|\boldsymbol{\hat{y}}_j - \boldsymbol{y}_j\|_2^2 \; .
\label{eq:L2}
\end{equation}

\paragraph{\bf Multivariate negative log-likelihood loss}
Predicting aleatoric uncertainty assumes a parametric noise distribution with a likelihood function. We then optimise the likelihood of the observed data as a function of the input and the distribution's parameters, using a negative log-likelihood (NLL) cost function \cite{bishop2006pattern}. Following the literature \cite{kendall2017uncertainties},  we model aleatoric uncertainty on the reconstructed pixel with a $K$-variate Normal distribution centered at the predicted value $\boldsymbol{\hat{y}}_j $ and with positive definite covariance matrix $\boldsymbol{\Sigma}$:

\begin{equation}
\mathcal{N}(\boldsymbol{y}_j|\boldsymbol{\hat{y}}_j, \boldsymbol{\Sigma}) = \frac{1}{\sqrt{|\boldsymbol{\Sigma|}}(2\pi)^\frac{K}{2}}  exp\left(-\frac{1}{2} \|\boldsymbol{\hat{y}}_j - \boldsymbol{y}_j \|_M \right) \;,
\label{eq:multi_Normal}
\end{equation}

with $\|.\|_M $ the Mahalanobis distance, defined as:
\begin{equation}
\|\boldsymbol{\hat{y}}_j - \boldsymbol{y}_j \|_M  = (\boldsymbol{\hat{y}}_j - \boldsymbol{y}_j )^T \boldsymbol{\Sigma}^{-1}(\boldsymbol{\hat{y}}_j - \boldsymbol{y}_j  ) \;.
\label{eq:Mahalanobis}
\end{equation}

Subsequently, the negative log likelihood loss writes as:

\begin{equation}
\mathcal{L}_{NLL}(\boldsymbol{y_j}|\boldsymbol{\hat{y_j}},\boldsymbol{\Sigma})  \varpropto \sum_{j=1}^{n} 
log(|\boldsymbol{\Sigma}_j|) + \|\boldsymbol{\hat{y}}_j - \boldsymbol{y}_j \|_M    \: .
\label{eq:NLL_multi_Gauss}
\end{equation}

Fitting a multivariate distribution raises the question of whether a full description of the covariance matrix should be pursued or if any structural constraints on $\Sigma$ are preferable. NLL optimization does become notoriously difficult when involving full covariance matrices \cite{skafte2019reliable, seitzer2021pitfalls}. 

\paragraph{\bf Diagonal covariance matrix} We define $\boldsymbol{\Sigma}$ as a diagonal matrix with diagonal elements $\boldsymbol{\sigma^2}= (\sigma_1^2, \cdots,\sigma_K^2) $. This greatly simplifies the inverse and determinant computations in \equaref{eq:NLL_multi_Gauss}. The diagonal model allows for different variance predictions per channel, which we experimentally find to be beneficial. However, cross-channel interactions in aleatoric predictions are not captured under this assumption, and such modelling is left for further research. To predict the variances, we set $C_{out}$ to $2\times K = 26$.
The diagonal entries of  $\boldsymbol{\Sigma}$ serve as aleatoric uncertainty prediction for the corresponding output channel:

\begin{equation}
    \boldsymbol{u_j} = [u_j^1,\cdots, u_j^K] = [\sigma_1^2, \cdots,\sigma_K^2] \:.
\end{equation}

\section{Experiments}

\subsection{Data} \label{sub:data}
We conduct our experiments on the SEN12MS-CR \cite{ebel2020multisensor} and SEN12MS-CR-TS \cite{ebel2022sen12ms} datasets for mono-temporal and multi-temporal cloud removal. Both are challenging image reconstruction benchmark datasets with about $50 \%$ cloud coverage over regions distributed across the whole planet and all seasons. The datasets contain ground range detected dual-polarization C-band $S1$ measurements as well as co-registered level-1C top-of-atmosphere reflectance $S2$ products, curated from Google Earth Engine \cite{gorelick2017google} and subsequently handled as documented in the two associated publications. 
The mono-temporal dataset contains $169$ regions, whereas SEN12MS-CR-TS focuses on a global subset of $53$ large areas. All regions of the datasets are utilized for training, validation and testing, with the respective splits as originally defined. Unless specified otherwise, experiments on SEN12MS-CR-TS are run on $T=3$ time points, which is a reasonable number of revisits for the cloud removal task and has been a prevalent choice in prior work \cite{Sarukkai_Jain_Uzkent_Ermon_2019, ebel2022sen12ms, sebastianelli2022plfm}. All data are of spatial dimensions $H=W=256$ px and we use the full spectrum of all $13$ optical bands. Analogous to preceding studies combining information of SAR and optical imagery \cite{eckardt2013removal, huang2015sar, meraner2020cloud, ebel2022sen12ms, xu2022glf} we use both Sentinel-1 and Sentinel-2 data to reconstruct images of the latter (i.e., $C_{S1}=2$, $C_{S2}=C_{out}=13$, and $C_{in}=C_{S1}+C_{S2}=15$). 
$S1$ data are preprocessed as in \cite{ebel2020multisensor, ebel2022sen12ms} and $S2$ pixel-values are divided by $1000$. 
Finally, binary cloud masks are calculated via s2cloudless \cite{Zupanc}---a lightweight and commonly deployed cloud detector \cite{s2cloudlessGEE, skakun2022cloud}. The cloud masks are used for sampling cloud-free target images at train time, statistical evaluations of results, and in prior work for losses that are cloud-sensitive \cite{meraner2020cloud}.

\subsection{Implementation details}

\paragraph{\bf Architectures} We train the proposed UnCRtainTS in its default setting with $n_e=1$ pre- and and $n_d=5$ post-aggregation MBConv blocks. The input convolution maps to $d_m=128$ channels, so that MBConv blocks map to $128 \rightarrow 256 \rightarrow 128$ channels with the default expansion factor $0.25$ in their Squeeze-Excitation layers. The L-TAE's parameters are kept to their default values $n_{head}=16$, and key dimension $d_k=4$. For mono-temporal considerations, we use the same architecture and simply discard the unnecessary L-TAE-based aggregation.  We compare our architecture against the baselines already evaluated on the SEN12MS-CR \cite{ebel2020multisensor} and SEN12MS-CR-TS \cite{ebel2022sen12ms} datasets.  We also evaluate the performance of U-TAE \cite{garnot2021panoptic} a state-of-the-art satellite image time series encoder, using the official implementation with minor adaptations to our task \footnote{github.com/VSainteuf/utae-paps}.  

\paragraph{\bf Training}
To assess the contribution of uncertainty modelling we train two variants: \textit{UnCRtainTS - no $\sigma$}, trained with L2 loss only, i.e., without uncertainty prediction, and \textit{UnCRtainTS} trained with the NLL loss of \equaref{eq:NLL_multi_Gauss} predicting uncertainties together with the reconstructed image. 
We use the ADAM optimizer \cite{kingma2014adam} with an initial learning rate of 0.001, at a batch size of 4 as in \cite{garnot2021panoptic}. All models are trained for 20 epochs with an exponential learning rate decay of 0.8, such that the rate decays by roughly one order of magnitude every 10 epochs. Models are evaluated on the validation split each epoch and the checkpoint with best validation loss is used for testing. 
\label{sub:training}

\paragraph{\bf Evaluation}\label{sub:evaluation} For image reconstruction performance, we report the Mean Absolute Error (MAE) or Root Mean Squared Error (RMSE) as well as Peak Signal-to-Noise Ratio (PSNR), Structural SIMilarity (SSIM) \cite{Wang_Bovik_Sheikh_Simoncelli_2004} and the Spectral Angle Mapper (SAM) metric \cite{kruse1993spectral}. We assess the quality of the uncertainty predictions via Uncertainty Calibration Error (UCE) \cite{guo2017calibration} 

\begin{equation}
UCE(e,u) = \sum_{p=1}^P \frac{N_p}{N} |e(B_p) - u(B_p)| \;,
\end{equation}\label{eq:UCE}

where $e(B_p)$ denotes the RMSE of $N_p$ pixel predictions in bin $B_p$, $P=20$ is the bin count and a bin's uncertainty $u(B_p)$ is given in terms of Root Mean Variance (RMV):

\begin{equation}
u(B_p) = \sqrt{\frac{1}{N_p } \sum_{j \in B_p}\frac{1}{K}  \sum_{k=1}^K u_j^k} \;.
\end{equation}\label{eq:RMV}

UCE quantifies the deviation between the predicted uncertainty and the empirical reconstruction error. Low UCE corresponds to well-calibrated uncertainties. We also report a patch-wise calibration metric termed UCE$_{im}$, where RMSE and RMV are spatio-spectrally averaged across all pixels of a given image before calculating calibration.

\subsection{UnCRtainTS} 
\label{subsection:eval_reconstruct}
In this section we show the experimental performance of our approach, both in terms of image reconstruction and aleatoric uncertainty prediction.

\paragraph{\bf Multi-temporal image reconstruction} We benchmark our method against established  heuristics and baselines of \cite{meraner2020cloud, Sarukkai_Jain_Uzkent_Ermon_2019, ebel2022sen12ms, garnot2021panoptic}.
We report the performance of these methods in \tabref{tab:sen12mscrts_L2}. UnCRtainTS sets a new state-of-the-art performance in terms of PSNR, SSIM, and SAM. Our architecture trained without uncertainty prediction (UnCRtainTS - no $\sigma$) scores second best on all those metrics and first in RMSE. This shows that our neural architecture alone  outperforms existing approaches, and uncertainty prediction further improves the reconstruction performance. Compared to U-TAE, the architecture improves by $1$pt SSIM while the uncertainty prediction increases the performance by another $0.7$pt. Note that uncertainty prediction has a slightly detrimental impact on RMSE performance ($-0.002$). This is in line with recent evidence that NLL optimization involves a trade-off between mean and variance estimate optimization that may hinder regression performance \cite{skafte2019reliable, seitzer2021pitfalls}. However this does not impact the image similarity metrics. Lastly, in terms of parameter efficiency, our model counts $0.5$M parameters. For comparison, the competitive U-TAE baseline \cite{garnot2021panoptic} which performs third-best consists of $1.2$M trainable weights, such that UnCRtainTS is relatively lightweight.

\begin{table}[!t]
\centering
\caption{\textbf{Multi-temporal image reconstruction experiment.} We evaluate models for $T=3$ inputs on SEN12MS-CR-TS benchmark. UnCRtainTS outperforms all learnable approaches on every metric, and performs best on all measures while predicting well calibrated uncertainties (bottom table).
    }
\scalebox{0.80}{
\begin{tabular}{@{}lllllllll@{}}
\toprule
Model        & $\downarrow$ RMSE & $\uparrow$ PSNR & $\uparrow$ SSIM & $\downarrow$ SAM  \\ \midrule
least cloudy & 0.079        & --- & 0.815 & 12.204 \\
DSen2-CR \cite{meraner2020cloud}   & 0.060       & 26.04 & 0.810 & 12.147 \\
STGAN \cite{Sarukkai_Jain_Uzkent_Ermon_2019}   & 0.057        & 25.42 & 0.818 & 12.548 \\
CR-TS Net \cite{ebel2022sen12ms}  & \textit{0.051}        & 26.68 & 0.836 & 10.657 \\
U-TAE \cite{garnot2021panoptic}   & \textit{0.051}        & 27.05 & 0.849 & 11.649 \\
UnCRtainTS - no $\sigma$ (ours)   & \textbf{0.049}        & \textit{27.23} & \textit{0.859} & \textit{10.168}\\ 
\textbf{UnCRtainTS} (ours)   & \textit{0.051}        & \textbf{27.84} & \textbf{0.866} & \textbf{10.160}\\ 
\bottomrule
\end{tabular}
 }
\newline
\vspace*{.05 cm}
\newline
\scalebox{0.80}{
\begin{tabular}{lll}
   \phantom{DSen2-CR \cite{meraner2020cloud} (least clo}   & UCE$_{im}$ &  UCE \\
  UnCRtainTS (ours)    & 0.010 &  0.007 \\
  \bottomrule
 \end{tabular}
}

    \label{tab:sen12mscrts_L2}
\end{table}

\begin{figure}[!h] 
    \includegraphics[width=1.0\linewidth]{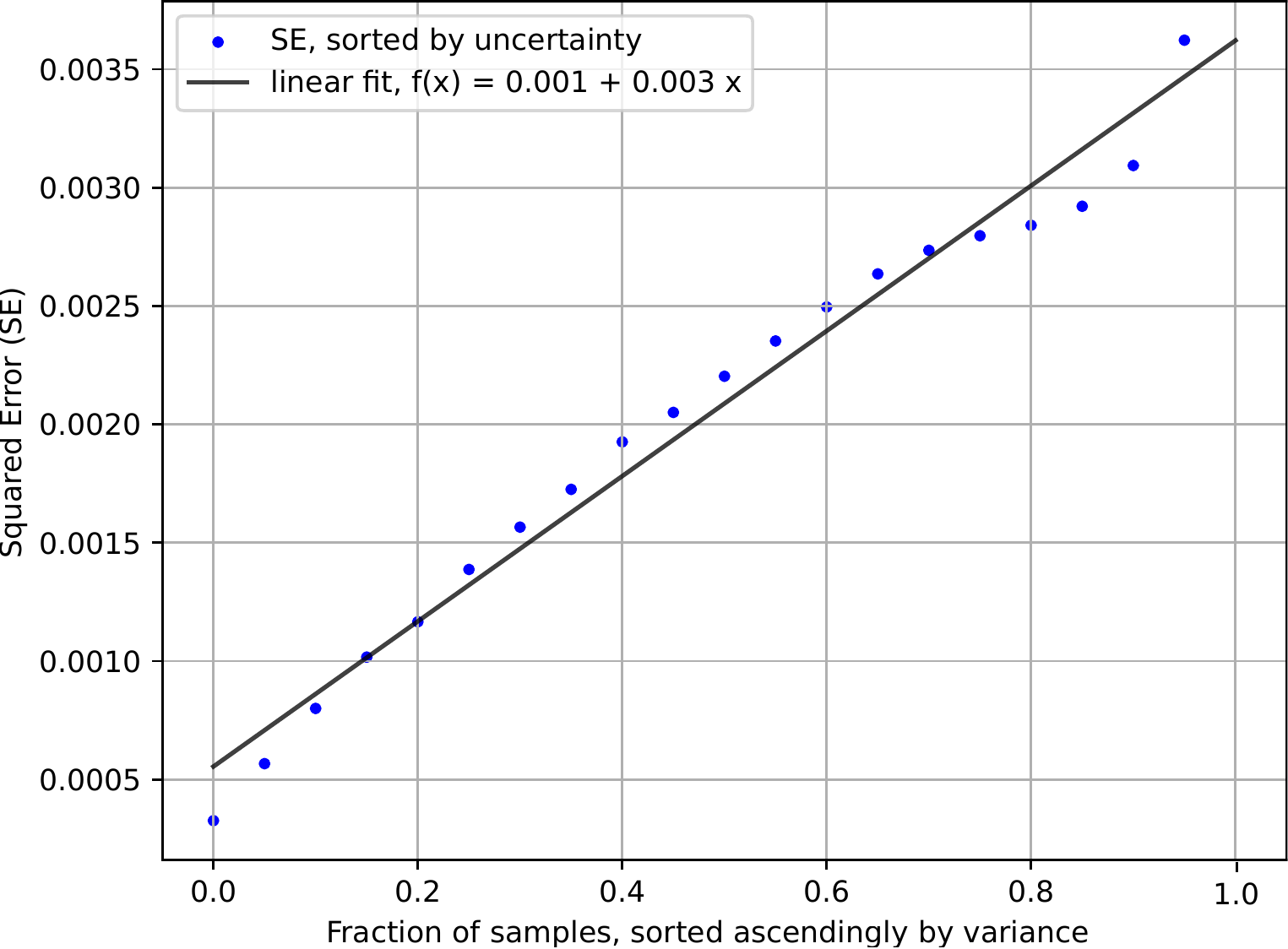}
    \caption{\textbf{Controlling error} on the test split by discarding top uncertain samples. 
    Discarding the top $50 \%$ of uncertain reconstructions almost halves prediction error, enabling risk management.
    }
    \label{fig:discard_all}
\end{figure}

\paragraph{\bf Aleatoric uncertainty prediction} We show the uncertainty calibration metrics of our method at image and 
pixel level in\tabref{tab:sen12mscrts_L2}. Those values should be compared to the test RMSE: at the pixel (resp. image) level the average error made on the reconstruction uncertainty is around $7$ (resp. $5$) times smaller than the average reconstruction error, showing satisfactory calibration. In other words, our method predicts uncertainty values that correlate well with the empirical reconstruction error. To demonstrate how uncertainty predictions can be useful in practice, we show how they allow filtering bad predictions.
We rank all reconstructed images of the test set sorted by increasing UCE$_{im}$ and accumulate squared errors from the least to the most uncertain samples.
The monotonous curve in \figref{fig:discard_all} displays a linear relation between error and uncertainty, such that error can be step-wise decreased by uncertainty-based filtering. In practice, this enables controlling risk in downstream applications on the restored satellite images.

\subsection{Architecture design}
To support the previous results and our architecture design choices, we systematically investigate UnCRtainTS' hyper-parameter sensitivity. Here, all model instances are trained with L2 loss only. Because UnCRtainTS operates on feature maps at full resolution, computational complexity is an important design criterion. In addition to its image reconstruction metrics, we report each model's number of trainable parameters and Floating Point Operations Per Second (GFLOPS), estimated via FAIR's \textit{fvcore} package \cite{fvcore}.

\begin{table}[!ht]
\centering
\caption{\textbf{Block setup.} Evaluation of the UnCRtainTS backbone for varying numbers of pre- and post-aggregation MBConv blocks. 
    }
\scalebox{0.7}{
\begin{threeparttable}
\begin{tabular}{@{}ccc|cc|ccccc@{}}
\toprule
\multirow{2}{*}
&  \multicolumn{2}{c|}{MBConv} & { params (k)} & {GFLOPS} & {$\downarrow$ RMSE} & {$\uparrow$ PSNR} & {$\uparrow$ SSIM} & {$\downarrow$ SAM}  \\ \cline{2-3} & $n_e$ & $n_d$ \\ \midrule
& 1 & 3 & 400 & 29.3 & 0.052        & 27.03 & 0.859 & 11.614 \\
& 1 & 4 & 483 & 34.0 & 0.050        & 27.00 & 0.851 & 11.771 \\
& 1 & 5 & 568 & 38.7 & \textit{0.049}        & \textit{27.23} & 0.859 & \textbf{10.168} \\ 
& 1 & 6 & 654 & 43.4 & 0.050        & \textbf{27.55} & \textit{0.860} & 10.471 \\
& 1 & 7 & 740 & 48.1 & 0.049        & 27.21 & 0.859 & \textit{10.300} \\ \midrule
& 0 & 5 & 483 & 24.6 & 0.052        & 26.97 & 0.853 & 11.002 \\
& 1 & 5 & 568 & 38.7 & \textit{0.049}        & \textit{27.23} & 0.859 & \textbf{10.168} \\ 
& 2 & 5 & 654 & 52.9 & \textbf{0.048}        & \textbf{27.55} & \textbf{0.864} & 10.641 \\
\bottomrule
\end{tabular}
\end{threeparttable}
}

    \label{tab:more_mbconv_L2}
\end{table}

\paragraph{\bf Spatial processing}
We explore the influence of the number of MBConv blocks before ($n_e$) and after ($n_d$) temporal aggregation in \tabref{tab:more_mbconv_L2}. 
Using $n_e=2$ blocks in the encoder instead of one, brings a $0.5$pt increase in SSIM, while the performance gain is marginal on the three other metrics. More pressingly, due to the parallel processing of the input sequence of feature maps, this setup incurs the highest computational complexity of 52.9 GFLOPS. 
In terms of post-aggregation blocks, performance peaks around $5-6$ modules, with $5$ modules being best on one metric and a close second on two more. 
For these reasons we choose $n_e=1$ pre and $n_d=5$ post aggregation blocks as default configuration. We also note that the ($n_e=0$) model performs competitively while being very lightweight and directly aggregating the input features. Indeed, it performs comparable to the U-TAE baseline. This secondary result shows that competitive performance can be obtained with very light architectures.

\begin{table}[!ht]
\centering
\caption{\textbf{Head count.} Quantitative evaluation of the UnCRtainTS backbone with varying number of self-attention heads.}
\scalebox{0.75}{
\begin{tabular}{@{}cc|cc|cccccc@{}}
\toprule
& $n_{head}$  &  params (k) & GFLOPS & {$\downarrow$ RMSE} & {$\uparrow$ PSNR} & {$\uparrow$ SSIM} & {$\downarrow$ SAM} \\ \midrule
& 1 & 556 & 38.7 & \textbf{0.049}        & \textbf{27.56} & 0.856 & 10.497 \\ 
& 4 & 559 & 38.7 & 0.052        & \textit{27.40} & 0.856 & 10.825 \\
& 8 & 563 & 38.7 & 0.051        & 27.00 & 0.851 & 11.131 \\ 
& 16 & 568 & 38.7 & \textbf{0.049}       & 27.23 & \textit{0.859} & \textbf{10.168} \\ 
& 32 & 588 & 38.8 & 0.051        & 27.12 & \textbf{0.861} & \textit{10.245} \\ 
& 64 & 621 & 38.9 & 0.051        & 27.24 & 0.858 & 11.054 \\ 
\bottomrule
\end{tabular}
}

    \label{tab:more_heads}
\end{table}

\paragraph{\bf Temporal aggregation} Second, we explore the effect of the number of attention heads on the reconstruction quality.
\tabref{tab:more_heads} shows that performances are closeby and differences in computational costs are negligible. 
We opt for $16$ heads, in line with the literature \cite{garnot2021panoptic}.

\paragraph{\bf Mono-temporal image reconstruction} 
To validate our resolution-preserving network design, we re-train and evaluate UnCRtainTS on the mono-temporal SEN12MS-CR dataset for cloud removal. That is, we consider the special case of $T=1$ to investigate the model's spatio-spectral restoration qualities and benchmark against the competitive baselines of \cite{Enomoto_Sakurada_Wang_Fukui_Matsuoka_Nakamura_Kawaguchi_2017, grohnfeldt2018conditional, Bermudez_Happ_Oliveira_Feitosa_2018, pan2020cloud, gao2020cloud, meraner2020cloud, xu2022glf}. Albeit being primarily designed for time series cloud removal, UnCRtainTS achieves best performances on all metrics except for SSIM, where it ranks second best following the recently published mono-temporal vision transformer architecture of \cite{xu2022glf}. The competitive performance achieved by the spatial encoding part of our architecture supports our choice of relying on MBConv blocks operating on full resolution feature maps.

\begin{table}[!h]
\centering
\caption{\textbf{Mono-temporal image reconstruction experiment.} Evaluation of models for $T=1$ inputs on the SEN12MS-CR benchmark. UnCRtainTS is best on all metrics except SSIM, where it is second following the recent vision transformer of \cite{xu2022glf}.
    }
\scalebox{0.80}{
    \centering
    \begin{tabular}{lllll}
    \toprule
        Method & $\downarrow$ MAE & $\uparrow$ PSNR & $\uparrow$ SSIM & $\downarrow$ SAM\\ \hline
        McGAN \cite{Enomoto_Sakurada_Wang_Fukui_Matsuoka_Nakamura_Kawaguchi_2017} & 0.048 & 25.14 & 0.744 & 15.676 \\ 
        SAR-Opt-cGAN \cite{grohnfeldt2018conditional} & 0.043 & 25.59 & 0.764 & 15.494 \\ 
        SAR2OPT \cite{Bermudez_Happ_Oliveira_Feitosa_2018} & 0.042 & 25.87 & 0.793 & 14.788 \\
        SpA GAN \cite{pan2020cloud} & 0.045 & 24.78 & 0.754 & 18.085 \\ 
        Simulation-Fusion GAN \cite{gao2020cloud} & 0.045 & 24.73 & 0.701 & 16.633 \\ 
        DSen2-CR \cite{meraner2020cloud} & 0.031 & 27.76 & 0.874 & 9.472 \\ 
        GLF-CR \cite{xu2022glf} & \textit{0.028} & \textit{28.64} & \textbf{0.885} & \textit{8.981} \\
        UnCRtainTS (ours) & \textbf{0.027} & \textbf{28.90} & \textit{0.880} & \textbf{8.320} \\  
 
        \bottomrule
    \end{tabular}
}

    \label{tab:sen12mscr}
\end{table}

\subsection{Uncertainty Modelling}

In this section, we provide additional experiments and ablations on the uncertainty prediction part of our method.

\begin{table}[!ht]
\caption{\textbf{Uncertainty models.} Evaluation of different uncertainty models and of two ensembles of $5$ UnCRtainTS instances (bottom), with and without SAR measurements as auxiliary input data.
    }
\centering
\scalebox{0.70}{
\begin{tabular}{c|cccc|cc}
\toprule
  model &  {$\downarrow$ RMSE} & {$\uparrow$ PSNR} & {$\uparrow$ SSIM} & {$\downarrow$ SAM} & $\downarrow$ UCE$_{im}$ & $\downarrow$ UCE \\ \midrule 
 
isotropic $\Sigma$ & 0.053 & 26.74 & 0.842 & 11.77 & 0.029 & 0.023 \\
UnCRtainTS  & 0.051 & 27.84 & 0.866 & \textbf{10.16} & \textbf{0.010} & \textit{0.007}  \\
\midrule
ensemble               & \textit{0.049} & \textbf{28.19} & \textbf{0.872} & \textit{10.18} & \textit{0.012} & \textbf{0.002} \\
ensemble$_{no SAR}$      & \textbf{0.048} & \textit{27.97} & \textit{0.869} & 10.76 & 0.018 & 0.014 \\
  \bottomrule
\end{tabular}
}

    \label{tab:cov_models}
\end{table}

\paragraph{\bf Comparison of covariance models} UnCRtainTS predicts aleatoric uncertainties using a diagonal covariance model, enabling different uncertainty predictions across channels. Here, this choice is compared to the simpler option of an isotropic covariance model. In the isotropic setting, we model the covariance matrix as $\boldsymbol{\Sigma}=\sigma^2 \boldsymbol{I_K}$ where $\sigma^2$ is scalar and $\boldsymbol{I_K}$ the $K$-dimensional identity matrix. This model assumes that the aleatoric uncertainty across channels can be described with a single value. We compare the performance of those two methods in \tabref{tab:cov_models}. The diagonal matrix model is best overall, outperforming on all metrics. 
These results clearly demonstrate that uncertainty prediction for satellite image reconstruction requires channel-specific uncertainty predictions. Indeed, modeling a diagonal covariance matrix over a simplistic isotropic description entails a three-fold reduction of the final uncertainty calibration error.

\paragraph{\bf Combined epistemic and aleatoric modelling} To give a full picture of uncertainty, we complement aleatoric uncertainty modelling with epistemic uncertainty estimation. We re-train the diagonal model with different weight initializations and samples of training batches to obtain a deep ensemble of $M=5$ member networks \cite{lakshminarayanan2017simple}. The members' reconstructions and uncertainty predictions are averaged via:
\begin{equation}
\hat{\boldsymbol{y}}^{M} = \frac{1}{M} \sum_{m=1}^{M} \hat{\boldsymbol{y}}^{m} 
\end{equation}\label{eq:ensemble_y}
\begin{equation}
(\sigma^{M})^2 = \frac{1}{M} \sum_{m=1}^{M} (\sigma^m)^2 + \frac{1}{M} \sum_{m=1}^{M} (\hat{\boldsymbol{y}}^{m})^2 - (\hat{\boldsymbol{y}}^{M})^2
\end{equation}\label{eq:ensemble_sigma}
to obtain the ensemble reconstruction $\hat{\boldsymbol{y}}^{M}$ and total uncertainty $(\sigma^{M})^2$. 
As shown on \tabref{tab:cov_models}, the $5$-member ensemble achieves the best reconstruction performances overall.
The full ensemble also achieves the best pixel-based calibration at $0.002$ UCE, 
Deep ensembles come at a computational cost both at training and inference time, but can prove valuable for the integration in downstream applications.

\begin{table}[!ht]
\caption{\textbf{Repeated Measures.} Evaluation of our ensemble of UnCRtainTS models with varying numbers of input time points.
    }
\centering
\scalebox{0.68}{
\begin{tabular}{@{}ll|llll|llll@{}}
\toprule
& input length T & { $\downarrow$ RMSE} & {$\uparrow$ PSNR} & {$\uparrow$ SSIM} & {$\downarrow$ SAM} & $\downarrow$ UCE$_{im}$ & $\downarrow$ UCE \\ \midrule
& 2 & 0.051 & 27.78 & 0.861 & 10.86 & 0.012 & 0.004 \\
& 3 & \textit{0.049} & \textit{28.19} & \textit{0.872} & \textit{10.18} & 0.012 & \textit{0.002} \\
& 4 & \textbf{0.047} & \textbf{28.41} & \textbf{0.875} & \textbf{9.99} & 0.013 & \textbf{0.001} \\
\bottomrule
\end{tabular}
}
    \label{tab:more_time_uncertainty}
\end{table}

\paragraph{\bf Uncertainty vs. sequence length} To evaluate the effect of the number of input time points $T$ on performances, we perform inference with the UnCRtainTS ensemble on input time series of lengths $T=2,3,4$.
\tabref{tab:more_time_uncertainty} shows that longer sequences help achieve both better image reconstruction quality and uncertainty calibration. 
This confirms the intuition that longer sequences, where additional samples are likely cloud-free, facilitate the restoration task and provide growing evidence for better calibration.  \tabref{tab:more_time_uncertainty} also underlines that the $T=3$ case considered in the main experiments makes for a challenging setting.

\paragraph{\bf SAR reduces uncertainty} We obtain a second ensemble trained without using SAR as auxiliary inputs, to explore the benefits of radar data. We show its performance on the bottom row of \tabref{tab:cov_models}. The single-sensor ensemble achieves a considerably higher UCE at both image and pixel level. This suggests that the additional information contained in the SAR inputs is beneficial to improve the trustworthiness of the reconstructions.

\begin{figure*}[!htb]
  \centering
  \begin{subfigure}[b]{0.17\linewidth}
    \includegraphics[width=\linewidth]{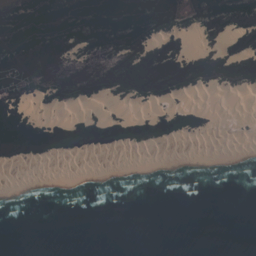}
  \end{subfigure}
  \begin{subfigure}[b]{0.17\linewidth}
    \includegraphics[width=\linewidth]{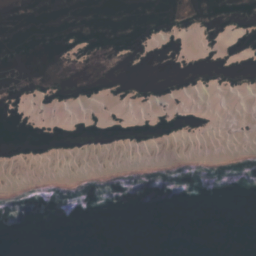}
  \end{subfigure}
  \begin{subfigure}[b]{0.17\linewidth}
    \includegraphics[width=\linewidth]{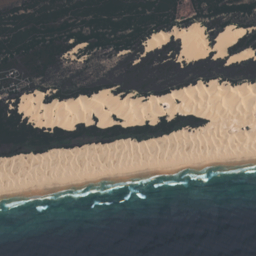}
  \end{subfigure}
  \begin{subfigure}[b]{0.17\linewidth}
    \includegraphics[width=\linewidth]{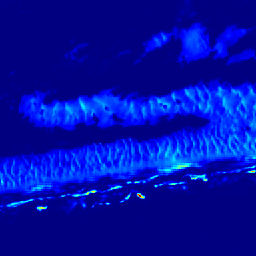}
  \end{subfigure}
  \begin{subfigure}[b]{0.17\linewidth}
    \includegraphics[width=\linewidth]{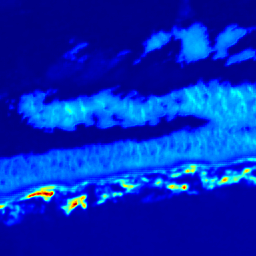}
  \end{subfigure}
  
  \begin{subfigure}[b]{0.17\linewidth}
    \includegraphics[width=\linewidth]{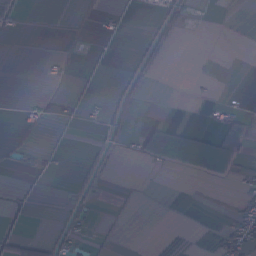}
  \end{subfigure}   
  \begin{subfigure}[b]{0.17\linewidth}
    \includegraphics[width=\linewidth]{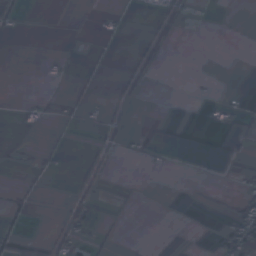}
  \end{subfigure}
  \begin{subfigure}[b]{0.17\linewidth}
    \includegraphics[width=\linewidth]{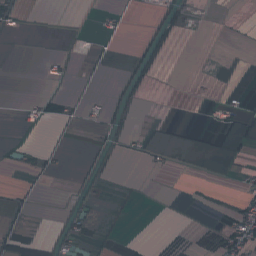}
  \end{subfigure}   
  \begin{subfigure}[b]{0.17\linewidth}
    \includegraphics[width=\linewidth]{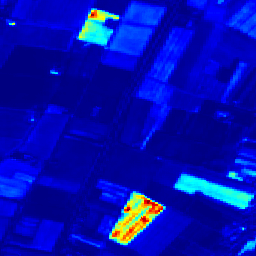}
  \end{subfigure}
  \begin{subfigure}[b]{0.17\linewidth}
    \includegraphics[width=\linewidth]{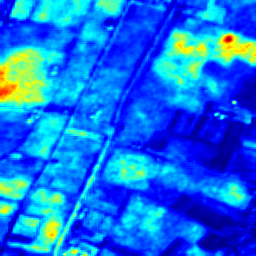}
  \end{subfigure}
  
  \begin{subfigure}[b]{0.17\linewidth}
    \includegraphics[width=\linewidth]{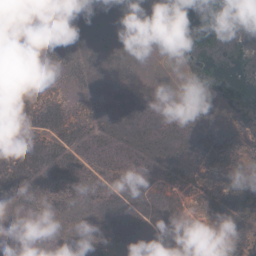}
  \end{subfigure}
  \begin{subfigure}[b]{0.17\linewidth}
    \includegraphics[width=\linewidth]{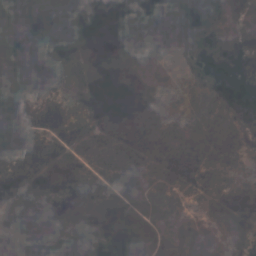}
  \end{subfigure}
  \begin{subfigure}[b]{0.17\linewidth}
    \includegraphics[width=\linewidth]{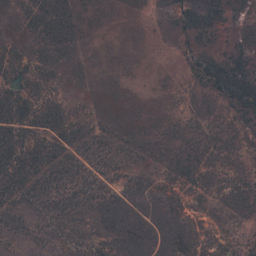}
  \end{subfigure}
  \begin{subfigure}[b]{0.17\linewidth}
    \includegraphics[width=\linewidth]{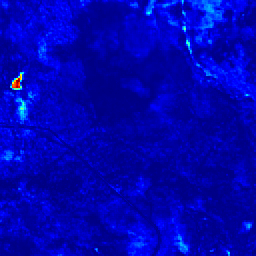}
  \end{subfigure}
  \begin{subfigure}[b]{0.17\linewidth}
    \includegraphics[width=\linewidth]{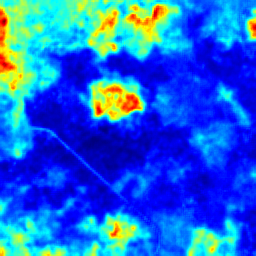}
  \end{subfigure}
   
  \begin{subfigure}[b]{0.17\linewidth}
    \includegraphics[width=\linewidth]{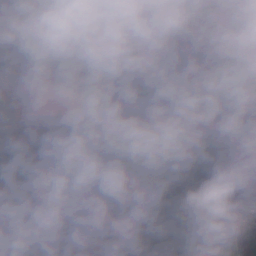}
    \caption{Cloudy input.}
  \end{subfigure}
  \begin{subfigure}[b]{0.17\linewidth}
    \includegraphics[width=\linewidth]{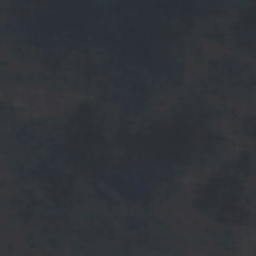}
    \caption{Prediction.}
  \end{subfigure}
  \begin{subfigure}[b]{0.17\linewidth}
    \includegraphics[width=\linewidth]{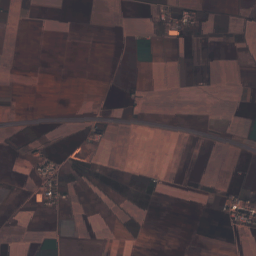}
    \caption{Target.}
  \end{subfigure}
  \begin{subfigure}[b]{0.17\linewidth}
    \includegraphics[width=\linewidth]{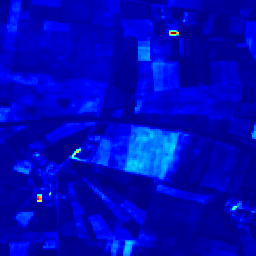}
    \caption{Error map.}
  \end{subfigure}   
  \begin{subfigure}[b]{0.17\linewidth}
    \includegraphics[width=\linewidth]{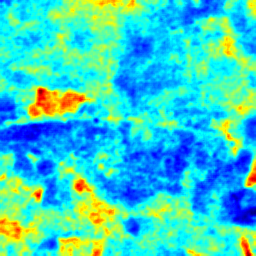}
    \caption{Uncertainty map.}
  \end{subfigure}
  
  \caption{\textbf{Exemplary images.} Detail view on exemplary satellite images and predictions by UnCRtainTS with a diagonal covariance matrix model. Rows: Four different samples from the test split. The illustrated cases show mild atmospheric distortions, semi-transparent haze, partly dense cloud coverage and cloud coverage with no visibility at all. Columns: The input sequence's least-cloudy image ($T=3$), UnCRtainTS' image reconstruction, the clear-view target image, the map of squared error residuals as well as the map of UnCRtainTS' variance predictions. Note the model's sensitivity to transients captured in the input time series, such as the ocean's white wash, changing crops as well as clouds and cloud shadow. UnCRtainTS captures these changing circumstances as data-inherent, aleatoric uncertainty. 
  }
  \label{fig:bunch_of_plots}
\end{figure*}

\paragraph{\bf Qualitative results} Complementary to the quantitative measures, \figref{fig:bunch_of_plots} shows UnCRtainTS' image restorations and uncertainty maps across varying levels of cloud coverage. Of particular interest is the uncertainty predictions not only being sensitive to clouds and cloud shadows, but also capturing other dynamics such waves breaking on a shore or the coloring of maturing crops. UnCRtainTS attends to differences in the input time series---not entirely unlike sequence-based cloud detectors explicitly designed for spotting transients across repeated measures \cite{MAJA}---and then, due to their temporary nature, attributes them an elevated aleatoric uncertainty.

\section{Conclusion}

We introduced UnCRtainTS, a novel method for combining uncertainty quantification with cloud removal from optical satellite image time series. While prior contributions applied uncertainty prediction in biomedical imaging or to univariate remote sensing downstream applications, our work is the first to investigate multivariate uncertainty quantification for multispectral satellite image reconstruction. UnCRtainTS features an attention-based neural architecture that outperforms all competitors benchmarked on the satellite image reconstruction task. Our proposed method includes a formulation of aleatoric uncertainty prediction for image reconstruction based on diagonal covariance matrices, as well as an estimation of epistemic uncertainty via deep ensembles. 
The conducted experiments show that both of our contributions, the new architecture combined with uncertainty quantification, set a new state-of-the-art image reconstruction performance on SEN12MS-CR-TS. Finally, the outcomes highlight how our well-calibrated uncertainties can effectively serve as a measure to control reconstruction quality and help integration in risk-sensitive downstream applications.
Our results encourage further explorations of more complex multivariate uncertainty models for image reconstructions. 
Our code is provided at \small{\url{https://patrickTUM.github.io/cloud_removal/}}.

\noindent
\paragraph{\bf Acknowledgements} \small{This work is jointly supported by the Federal Ministry for Economic Affairs and Energy of Germany in the project “AI4Sentinels– Deep Learning for the Enrichment of Sentinel Satellite Imagery” (FKZ50EE1910), by the German Federal Ministry of Education and Research (BMBF) in the framework "AI4EO -- Artificial Intelligence for Earth Observation: Reasoning, Uncertainties, Ethics and Beyond" (01DD20001) and by the German Federal Ministry of Economics and Technology in the framework of the "national center of excellence ML4Earth" (50EE2201C).}

\clearpage

%%%%%%%%% REFERENCES
{\small
\bibliographystyle{ieee_fullname}
\bibliography{egbib}
}

\end{document}